\documentclass[letterpaper, 10 pt, conference]{ieeeconf}  

\IEEEoverridecommandlockouts  

\usepackage{graphics} 
\usepackage{epsfig} 
\usepackage{mathptmx} 
\usepackage{times} 
\usepackage{amsmath} 
\usepackage{amssymb}  

\usepackage{amsfonts}
\usepackage{algorithmic}
\usepackage{algorithm}
\usepackage{array}
\usepackage[caption=false,font=normalsize,labelfont=sf,textfont=sf]{subfig}
\usepackage{textcomp}
\usepackage{stfloats}
\usepackage{verbatim}
\usepackage{graphicx}
\usepackage{cite}
\usepackage{url}

\usepackage{multirow} 
\usepackage{color} 

\title{\LARGE \bf
Robust simultaneous UWB-anchor calibration and robot localization for emergency situations
}

\author{Xinghua Liu and Ming Cao 
\thanks{*Liu's work was supported in part by the China Scholarship Council. The work of Cao was supported in part by the Dutch Research Council (NWO) through its Vici project 19902.}
\thanks{Both authors are with the Institute of Engineering and Technology (ENTEG) at the Faculty of Science and Engineering, University of Groningen, 9747 AG Groningen, the Netherlands. 
        {\tt\small \{xinghua.liu, m.cao\}@rug.nl}}%
}

\usepackage{fancyhdr}
\fancypagestyle{IEEEcopyright}{
    \fancyhf{} 

    \fancyfoot[C]{\footnotesize
    © 2025 IEEE. Personal use of this material is permitted. Permission from IEEE must be obtained for all other uses, in any current or future media, including reprinting/republishing this material for advertising or promotional purposes, creating new collective works, for resale or redistribution to servers or lists, or reuse of any copyrighted component of this work in other works. DOI: 10.XXXX/XXXXX}
}

\usepackage[pscoord]{eso-pic}
\newcommand{\copyrightnotice}{
  \AddToShipoutPictureFG*{
    \put(55,15){%
      \parbox[b]{\textwidth}{%
        \centering
        \footnotesize
        \textcopyright~ 2025 IEEE. Personal use of this material is permitted. Permission from IEEE must be obtained for all other uses, in any current or future media, including reprinting/republishing this material for advertising or promotional purposes, creating new collective works, for resale or redistribution to servers or lists, or reuse of any copyrighted component of this work in other works. DOI: 10.XXXX/XXXXX
      }
    }
  }
\AddToShipoutPicture*{
  \put(55,\paperheight-40){%
    \parbox[b]{\textwidth}{%
      \footnotesize
      Accepted Article. IEEE copyrighted material.
}}}
}

\begin{document}

\maketitle

\copyrightnotice
\pagestyle{IEEEcopyright}

\begin{abstract}
In this work, we propose a factor graph optimization (FGO) framework to simultaneously solve the calibration problem for Ultra-WideBand (UWB) anchors and the robot localization problem. Calibrating UWB anchors manually can be time-consuming and even impossible in emergencies or those situations without special calibration tools. Therefore, automatic estimation of the anchor positions becomes a necessity. The proposed method enables the creation of a soft sensor providing the position information of the anchors in a UWB network. This soft sensor requires only UWB and LiDAR measurements measured from a moving robot. The proposed FGO framework is suitable for the calibration of an extendable large UWB network. Moreover, the anchor calibration problem and robot localization problem can be solved simult.aneously, which saves time for UWB network deployment. The proposed framework also helps to avoid artificial errors in the UWB-anchor position estimation and improves the accuracy and robustness of the robot-pose. The experimental results of the robot localization using LiDAR and a UWB network in a 3D environment are discussed, demonstrating the performance of the proposed method. More specifically, the anchor calibration problem with four anchors and the robot localization problem can be solved simultaneously and automatically within 30 seconds by the proposed framework. The supplementary video and codes can be accessed via \url{https://github.com/LiuxhRobotAI/Simultaneous_calibration_localization}.
\end{abstract}

\section{INTRODUCTION}
A challenging problem in indoor localization with wireless sensor networks, in particular, Ultra-WideBand (UWB) networks, is to determine the position of anchors automatically (namely \emph{calibration} of UWB anchors), especially in emergency situations where the demand for network deployment methods is high. The growing application of indoor robot localization, e.g. indoor parking lots, cargo automation management in automatic factories, warehouses, and fire rescue\cite{ridolfi2021self, jang_survey_2023}, gives rise to the additional sensing capability constraints when calibrating. Furthermore, the problem of simultaneous calibration or \emph{cooperative localization} for UWB anchors and localization of the robot position in real environments remains to be further studied.
\begin{figure*}[thpb]
      \centering
      \includegraphics[width=6in]{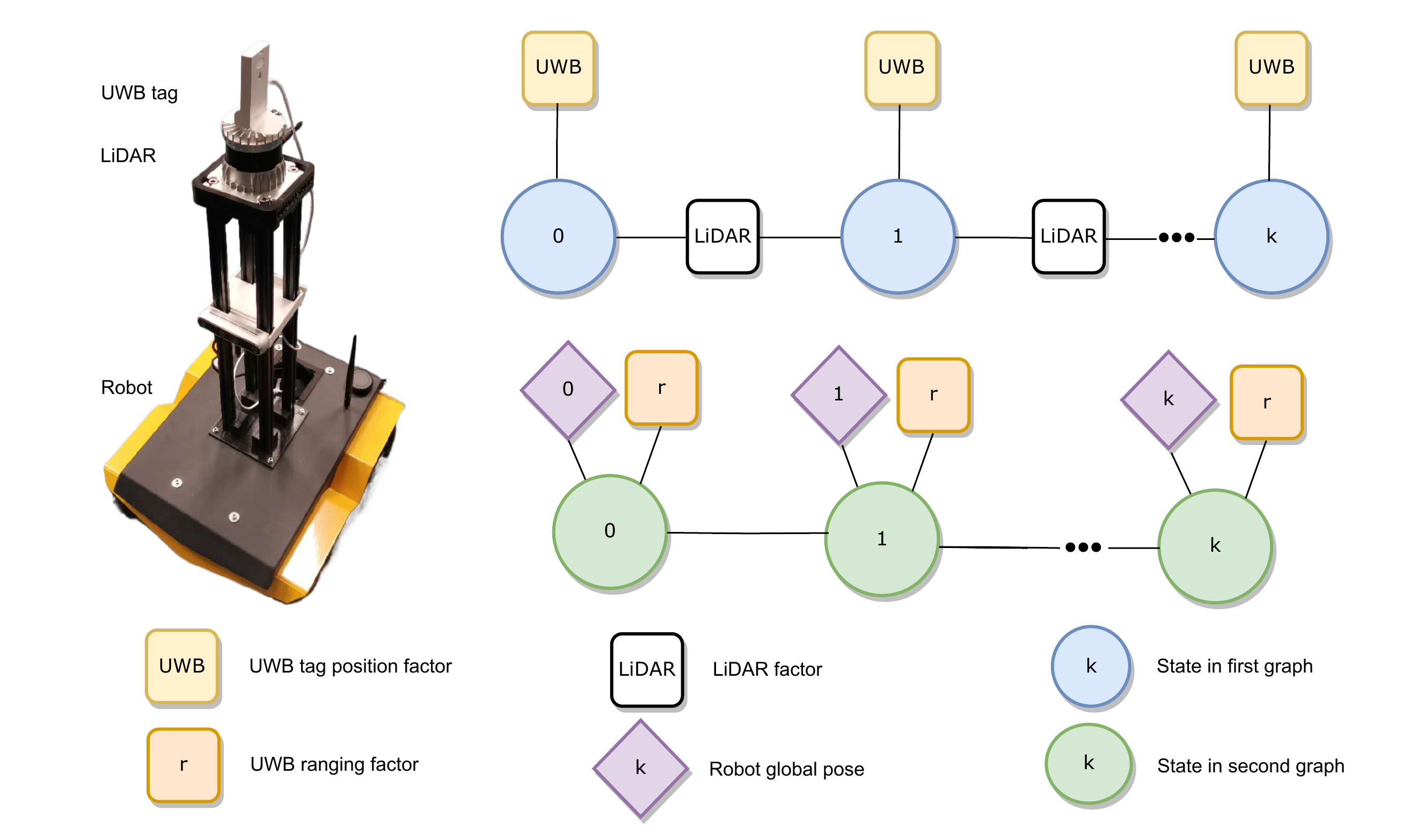}
      \caption{An illustration of the proposed framework for simultaneous anchor calibration and robot localization, and the robot platform in experiments. The robot is equipped with a UWB tag to receive the ranging measurements between UWB anchors and the robot, and a 3D LiDAR for LiDAR SLAM. The first graph demonstrates the robot localization problem with the large blue circle representing the state to be estimated including the global poses of the robot and the frame transformation. The second graph describes a UWB anchor calibration problem utilizing the global poses estimated in the first graph and UWB ranging measurements.}
      \label{pose_graph_structure}
\end{figure*}

Calibration of a UWB network is indispensable in certain cases since manually measuring the position requires some special calibration tools, like a rangefinder or a total station. To find the positions of UWB anchors, one can design a soft sensor or a virtual sensor, which is a program processing several measurements together, to calculate the required positions, instead of the usage of a real hard sensor to measure them. A factor graph optimization-based approach is appropriate for the role of a soft sensor because the calibration of UWB anchors can be modeled as a maximum a-posteriori (MAP) estimation problem \cite{dellaert2012factor, Dellaert2017Factor}. The factors in the factor graph represent the posterior probability of the anchor positions concerning known distance measurements between UWB anchors and moving tags. Factor graph optimization (FGO) is well suited to solving complex estimation problems, such as the structure from motion and simultaneous localization and mapping (SLAM), and solving data fusion problems with multiple sensors \cite{qin2018vinsmono,qin2019general}. Yuan et al. \cite{yuan2019toa} present a factor graph-based framework and the regularized least square (RLS) for localization problems. However, the proposed framework only works in passive localization scenarios while robot localization with UWB is an active localization problem in general.

Similar to anchor calibration, factor graphs are also useful in simultaneous localization and mapping with LiDAR in robot localization. A well-known LiDAR SLAM approach named LOAM (\underline{L}iDAR \underline{O}dometry \underline{a}nd \underline{M}apping) is proposed by Zhang \cite{zhang2014loam}, and then implemented by Qin. et al. \cite{hkust_aloam} in 2019 reaching a real-time performance with a 3D LiDAR. As LOAM does not include loop closures or dynamic objectives detection, the high precision of pose estimation does not persist for a long time. There are many techniques to improve the accuracy of a LiDAR SLAM in dynamic environments \cite{Shan2018LeGOLOAM, xu2022fast, liu2023dlcslam, tosi2024nerfs}. These LiDAR SLAM algorithms significantly improve the localization accuracy of robots moving within local areas. However, as they extract only relative poses of a robot from the LiDAR point cloud, they are prone to drift over extended distances due to the absence of global information.

There are three main methods for UWB localization problems: triangulation or trilateration, scene analysis, and proximity \cite{ridolfi2021self}; it is not surprising that trilateration is one of the most focused approaches to UWB calibration as UWB is often used as a range-based technology. A popular UWB manufacturer provides a procedure based on trilateration for the auto-positioning of anchors \cite{jin2021auto}. This is a \emph{self-calibration} technology only applicable to a UWB network with four anchors at the same height. This scheme with such restrictions provides only an accuracy of around 0.5m (and greater than 1 meter in some worst scenarios for the positioning error of the anchors). Ridolfi et al. propose a neural network to study, detect, and correct measurement errors in a self-calibration problem of a UWB network in \cite{ridolfi2021uwb}. Although their self-calibration algorithm works in a realistic non-line-of-sight (NLOS) environment, the consideration of adaptive physical settings requires implementing an adaptive PHY-layer modulation, which increases the difficulty of using UWB for localization. Some researchers \cite{Corbalán2023SelfLocalization} suggest applying a multidimensional scaling (MDS) approach to locate the anchors, while MDS only works with UWB networks in which anchors can range each other. Kolakowski et al. propose a UWB anchor calibration approach with the assistance of a static LiDAR in \cite{kolakowski2022static}. However, the LiDAR is only used to localize the UWB anchors and should be placed manually in a restricted area of the deployment environment to reach satisfactory efficiency. 

Simultaneous calibration and localization is a challenging topic to study, in which anchor calibration problems and robot localization problems need to be solved with the same framework. Shi et al. \cite{shi2019anchor} propose a calibration method strategy that relies on an Error-State Kalman Filter (ESKF), which utilizes low-cost IMU data and UWB range data. The proposed algorithm estimates the locations of the anchors and a tag automatically by freely moving the tag in the environment. However, their method relies heavily on the continuous reading of IMU and UWB. If the reading of the UWB anchor or IMU is interrupted for a while, the calibration process cannot be completed; to make it even worse, the positioning process cannot be carried out, even if the readings return to normal shortly after the interruption. Hence, the robustness cannot be guaranteed. Similar work is proposed by Qi et al. \cite{Qi2024Calibration}, in which they calibrate and compensate anchor positions with an IMU and UWB network but with a derivative unscented Kalman filter, which also requires distance constraints between anchors. They also use the second-order nonlinear terms of anchor position error to improve the calibration accuracy in their tight integration algorithm of IMU and UWB, which makes it more sensitive to the interruption of the sensor reading.

In this paper, we propose a robust factor graph optimization-based framework shown in Fig. \ref{pose_graph_structure} to solve the calibration problem for UWB anchors and locate a robot simultaneously in a real indoor environment. The proposed method enables the creation of a soft sensor that utilizes the position estimations of a moving robot equipped with a LiDAR and a UWB tag, and UWB ranging measurements between the tag and anchors to calculate the position of UWB anchors automatically.
The calibration problem and robot localization problem can be solved simultaneously, which saves time for UWB network deployment. Moreover, the proposed method is applicable for larger and extendable UWB networks as the only needed information to calibrate a new anchor is the ranging measurements between the moving tag and this new anchor.

The main contributions of this work are summarized as follows:
\begin{enumerate}
    \item A robust simultaneous UWB-anchor calibration and robot localization framework is proposed. The proposed framework is based on factor graph optimization technology, which enables the creation of a soft sensor providing the position information of anchors in an extendable UWB network for emergencies. The proposed framework utilizes a factor graph processing LiDAR data and robot position data from UWB for real-time data fusion, even if the reading of the UWB anchors or the LiDAR is interrupted for some time. The UWB ranging measurements are used to estimate the position of UWB anchors with nonlinear optimization in another factor graph.
     \item The transformation between the fixed UWB frame and the Lidar initial frame is estimated stably within one of the factor graphs to allow online simultaneous calibration of UWB anchors and localization of a robot. The fixed UWB frame is a human-known frame, which is helpful for subsequent robot tasks.
    \item The performance in a real indoor environment for simultaneously solving the 3D robot localization problem and the UWB-anchor calibration problem has been discussed.
\end{enumerate}

The rest of this paper is organized as follows. In Section \ref{Related_Work}, some related works that are useful for the proposed framework are introduced. In Section \ref{Simultaneuos_calibration_Localization}, the details of the proposed factor graph-based optimization framework are presented. The simultaneous UWB-anchor calibration and robot localization problem is modeled. The transformation between the UWB frame and the Lidar initial frame is also discussed in this section. In Section \ref{Experiments}, some experimental results are given to demonstrate the validation of the proposed method. And then, in Section \ref{CONCLUSIONS}, the summarization and discussion of this paper are provided.

\section{Related Work} \label{Related_Work}
\subsection{Factor Graph Optimization}
The factor graph approach is motivated by Bayesian networks for hidden Markov models\cite{dellaert2012factor}. A factor graph includes nodes and edges representing the constraint functions or variables and the relationship between them. Factors in the graph represent functions on subsets of the variables, which can be formulated with measurements or constraints. One of the key advantages of factor graphs is their ability to model complex estimation problems, such as simultaneous localization and mapping in robotics.

Factor graph optimization is an effective and powerful approach based on factor graphs, and aims to maximize the posterior probability of variables of interest based on a given set of measurements. Many optimization problems in robotics \cite{wang2017ultra,guo2019ultra} can be modeled as linear or nonlinear least squares optimization problems, which can be visually represented in a graphical form, and thus can be solved by factor graph optimization with specific tools, such as Ceres Solver\cite{agarwal2022ceres_solver} and Theseus \cite{pineda2022theseus}.

Factor graph optimization has emerged as a powerful tool for solving robot localization problems. One SOTA implementation using factor graph optimization to solve a SLAM problem in robotics can be found in \cite{qin2019general}, which uses GPS factors and VIO (Visual inertial odometry) factors to estimate the global poses of a robot. However, this framework cannot reliably estimate a fixed transformation between the global frame (GPS frame or Earth fixed frame) and the local frame (VIO initial frame). Consequently, this framework is unsuitable for UWB-anchor calibration. Despite these drawbacks of the framework, we are motivated to explore an FGO-based framework for simultaneous anchor calibration and robot localization.

\subsection{Simultaneous Localization and Mapping with LiDAR} \label{LiDARodometry}
Simultaneous localization and mapping with LiDAR, also known as LiDAR odometry and mapping (LOAM) \cite{zhang2014loam} in certain cases, plays a crucial role in robot localization. One notable implementation of LOAM with real-time performance is A-LOAM \cite{hkust_aloam}. The main goal of a SLAM problem with LiDAR is to recover the LiDAR motion by utilizing the information of the feature points. LOAM does not include loop closures or dynamic object detection, so it cannot correct for drift over time. Despite its limitations, LOAM maintains high accuracy over a short duration in static environments. Researchers have observed similar conclusions in other LiDAR-based SLAM approaches, including the state-of-the-art LeGO-LOAM \cite{Shan2018LeGOLOAM}. Its accuracy is sufficient for the anchor calibration problem. LOAM performs well for UWB anchor calibration problems as the calibration process can be completed in a very short period.

\subsection{Calibration of UWB Anchors} \label{selfCalibrationAnchor}
To determine the position of anchors in a UWB network, one method is to calculate the position using the ranging measurement between anchors, which becomes a self-calibration problem of a UWB network. As the anchors are deployed at fixed positions, one can set some coordinates of some anchors directly to be constant to simplify the calibration problem, i.e., setting one anchor as the origin and setting the other one located in its x or y direction. A scheme provided by a manufacturer of the UWB module is to collect mutual ranges of UWB anchors by Bluetooth on a mobile device that communicates with the anchors\cite{Decawave2017MDEK1001}, and then these ranges are used to calculate the positions of anchors by trilateration. This self-calibration scheme only works for anchors in the same plane, which requires them to be at the same height and provides an accuracy greater than 1 meter for the positioning error of the anchors in worst scenarios \cite{jin2021auto}. To make it even worse, while self-calibrating, the anchors are unavailable for localizing the tags, and if the positions of the anchors change after self-calibration, another self-calibration procedure is required. However, the position estimation of this scheme can be used as an initial value for further improved calibration methods, and multiple self-calibrations can provide a more reliable estimation of the anchor position.

More practical scenarios arise where communication between anchors within the UWB network is hindered. In such cases, the only available measurement is the distance between tags and anchors. The key issue in these situations is estimating the coordinates of anchors and subsequently building a reliable map of these anchors, using only ranging information between tags and anchors. Moreover, in critical situations, such as emergencies, simultaneous UWB anchor calibration and robot localization becomes imperative. Solving these problems requires robust and innovative techniques that leverage the limited information available from tag-anchor distance measurements. 

\section{Simultaneous Calibration and Localization} \label{Simultaneuos_calibration_Localization}
We propose an innovative FGO-based framework to address the simultaneous UWB-anchor calibration and robot localization. Our approach aims to enhance the robustness and reliability of positioning systems in complex environments. The details of the proposed framework are described in this section.
\subsection{Factor on LiDAR measurements}
One factor in the factor graph is constructed and associated with the LiDAR measurements. We leverage a classical LOAM \cite{hkust_aloam} to provide local pose constraints and construct the corresponding factor. The robot frame is denoted as $\{R\}$ corresponding to the subscript or superscript $r$ in the parameters. The LiDAR odometry estimates the states of a robot in the real-time LiDAR frame $\{L\}$ and the pose transformations to the LiDAR initial frame (referred to as the map frame, denoted as $\{M\}$). Every two adjacent pose transformations $T^m_{l,k-1}$ and $T^m_{l,k}$ in the parameter set $T^m_r$ are used to construct a relative pose factor. Let the notation $f_l(T^g_r; T^m_l)$ represent the cost functions with respect to the variable set $T^g_r$ with the LiDAR observation set $T^m_l$, which emphasizes that a cost $f_{l,k}$ is a function of the global pose of the robot $T^g_{r,k}$ in the set $T^g_r$ with LiDAR observations $T^m_{l,k-1}$ and $T^m_{l,k}$ in the set $T^m_l$. These LiDAR observations $T^m_l$ are used to construct the following cost functions:
\begin{equation} \label{LiDAR_factor}
\begin{aligned}
& f_l(T^g_r;T^m_l) := \frac{1}{2} \sum_{k=1}^{N_l} \rho_{l,k} \left\{ \left\| f_{l,k} \left(T^g_{r,k},T^g_{r,k-1}; T^m_{l,k-1}, T^m_{l,k} \right) \right\|^2 \right\},\\
\end{aligned}
\end{equation}
\begin{equation}
\begin{aligned}
& f_{l,k} \left(T^g_{r,k},T^g_{r,k-1}; T^m_{l,k-1}, T^m_{l,k} \right) = T^g_{r,k} T^{r,k-1}_g -T^m_{l,k} T^{l,k-1}_m,\\
\end{aligned}
\end{equation}
where the set $T^g_r$ to be estimated includes $N_l$ robot global poses $T^g_{r,k}$ expressed in a global frame $\{G\}$ at the time instant $t_k$, $T^m_l$ is a pose set of the LiDAR odometry including $N_l$ pose transformations $T^m_{l,k}$ expressed in the map frame $\{M\}$, $f_{l,k}$ is a cost function with respect to the variable $T^g_{r,k}$ calculated with two observations of pose transformation from the LiDAR odometry in instant $t_{k-1}$ and $t_k$, and $\rho_{l,k}$ is a loss function for the LiDAR factor that can be the identity function, Huber loss function, and so on. We use the identity function as our loss function for the LiDAR factors since the LiDAR pose estimation is accurate enough during the period throughout the entire calibration process.

\subsection{Factor on position estimation with the UWB network}
A self-calibration procedure is performed to get rough initial positions of the anchors in the UWB network. The initialized anchors are then only used to localize the robot and provide rough positions $P^u$ for the robot. These robot positions estimated by the UWB network and the robot poses observed by the LiDAR odometry are used to construct the cost functions with respect to the robot global poses $T^g_r$. These cost functions are presented as follows:
\begin{equation} \label{UWB_position_factor}
\begin{aligned}
f_u({T^g_r;P^u,T^m_l}) :=& \frac{1}{2} \sum_{k=1}^{N_u} \rho_{u,k} \left\{ \left\|  f_{u,k} \left(T^g_{r,k}; p^u_{k}, h_l(T^m_{l,k}) \right) \right\|^2 \right\} \\
=& \frac{1}{2} \sum_{k=1}^{N_u} \rho_{u,k} \left\{ \left\| f_{u,k} \left( T^g_{r,k}; p^u_{k}, p^m_{k} \right) \right\|^2 \right\},\\
\end{aligned}
\end{equation}
\begin{equation}
\begin{aligned}
f_{u,k} \left( T^g_{r,k}; p^u_{k}, p^m_{k} \right) &=  p^u_{k}- T^g_{r,k} p^m_{k},
\end{aligned}
\end{equation}
where ${P^u}$ is a position set of the robot including $N_u$ position of the robot estimated by the UWB network expressed in the UWB frame $\{U\}$, $f_{u,k}$ is a cost function of $T^g_{r,k}$ constructed with the robot position observation $p^u_{k}$ from the UWB network and observation $p^m_{k}$ from the LiDAR odometry at the same time instant $t_k$, $h_l$ is a function extracting the robot position from a given robot pose, and the loss function $\rho_{u,k}$ is chosen to be the Huber loss, which strikes a balance between robustness and sensitivity, making it suitable for handling outliers and noisy data. To simplify, the transformed position $p^g_{k}$, which is derived from $[(p^g_{k})^T, 1]^T = T^g_{r,k} [(p^m_{k})^T, 1]^T$ is denoted by $T^g_{r,k} p^m_{k}$. In the remaining of the paper, similar simplifications in notation are used wherever no ambiguity is foreseen.

\subsection{Frame transformation and data fusion}
To express the pose of the robot in a human-known frame (a fixed frame that is comprehensible to humans, not only known by the robot), the frame transformation between the map frame and the global frame must be calculated. One of the human-known frames is the UWB frame determined by the UWB network because one can strategically design the placement of anchors to satisfy specific requirements relevant to subsequent robot tasks. Hence, we choose the fixed UWB frame $\{U\}$ as the global frame $\{G\}$ in the proposed framework. Next, we describe the estimation of the frame transformation between the map frame and the UWB frame within the same FGO-based framework. One can estimate the frame transformation $T^u_m$ by utilizing the LiDAR odometry observations $T^m_l$ and the robot position observations $P^u$ provided by the UWB network. The cost functions are given as follows:
\begin{equation} \label{global_transformation_factor}
\begin{aligned}
f_t(T^u_m;P^u,T^m_l) :=& \frac{1}{2} \sum_{k=1}^{N_u} \rho_{t,k} \left\{ \left\|  f_{t,k} \left(T^u_m; p^u_{k} , h_l(T^m_{l,k}) \right) \right\|^2 \right\} \\
=& \frac{1}{2} \sum_{k=1}^{N_u} \rho_{t,k} \left\{ \left\| f_{t,k} \left( T^u_m; p^u_{k}, p^m_{k} \right) \right\|^2 \right\},\\
\end{aligned}
\end{equation}
\begin{equation} 
\begin{aligned}
f_{t,k} \left( T^u_m; p^u_{k}, p^m_{k} \right)=& p^u_{k}-T^u_m p^m_{k},
\end{aligned}
\end{equation}
where ${T^u_m}$ is the frame transformation between $\{U\}$ and $\{M\}$ to be estimated, $f_{t,k}$ is the cost function of the frame transformation calculating the residual between the position $p^u_{k}$ and the observation $p^m_{k}$ after transforming by ${T^u_m}$ in instant $t_k$, and $N_u$ is the number of the robot positions from the UWB network. The loss function $\rho_{t,k}$ is chosen to be the Huber loss in the cost function.

With the transformation between the UWB frame and the LiDAR initial frame calculated stably, one can estimate the pose of the robot in the global frame using LiDAR measurements and UWB measurements, which is a data fusion process. Combining the cost functions above, the data fusion process can be described with the following nonlinear least squares problems:
\begin{equation} \label{fgo_cost_function}
\begin{aligned}
&\min _{\{T^g_r,T^u_m\}} \left\{f_l(T^g_r;T^m_l) + f_u(T^g_r;P^u,T^m_l) + f_t(T^u_m;P^u,T^m_l) \right\}. \\
\end{aligned}
\end{equation}
In this framework, one can get a robust position estimation of the robot, even if the reading of the UWB anchors or the LiDAR is interrupted for a while, which means that the position estimation process will continue with only one sensor reading and ensure an acceptable estimation accuracy during the reading interruption of another sensor. The position estimation will be improved and the data fusion process will be carried out after the readings return.

\subsection{Factor on UWB ranging and anchor calibration}
After estimating the global poses of the robot and the stable pose transformation between the UWB frame and the map frame, one can calibrate the UWB anchors with the ranging measurements now. We construct factors using the ranging measurements and the global position in the global poses to solve the calibration problem, which is modeled in the following form:
\begin{equation} \label{calibration_f}
\begin{aligned}
& \min _{X^u} \frac{1}{2} \sum_{k=1}^{N_r} \sum_{i \in{\mathrm{P_a}}}\rho_{r,k} \left\{ \|  f_{uwb}(X^u _ {i}; p^u_{k},r_{i,k}) \|^2 \right\},
\end{aligned}
\end{equation}
\begin{equation}
\begin{aligned}
f_{uwb}(X^u _ {i}; p^u_{k},r_{i,k})=\|X^u _ {i}- p^u_{k}\|-r_{i,k}, i \in{\mathrm{P_a}},
\end{aligned}
\end{equation}
where $X^u _ {i}$ is the position of the $i$-th anchor of the UWB network in the anchor position set $X^u$, $r_{i,k}$ is the corresponding rangings between the UWB tag carried on the robot and the UWB anchor in the anchor set $\mathrm{P_a}$, $N_r$ is the corresponding number of the ranging measurements, $f_{uwb}$ is the cost function with respect to the anchor position $X^u _ {i}$ incorporating distance measurement $r_{i,k}$ and robot position observation $p^u_{k}$, and the loss function $\rho_{r,k}$ is chosen to be the Huber loss here.

The FGO-based framework ensures a calibration scheme for an extendable UWB network as only the corresponding ranging measurement is required to estimate the new anchors in this framework. The proposed FGO-based framework includes two factor graphs for robot localization and UWB anchor calibration, respectively. The constructed graphs are solved by a graph optimization tool, Ceres Solve\cite{agarwal2022ceres_solver}, which provides an automatic differentiation procedure and can calculate the Jacobian. An illustration of the proposed pose graph structure is given in Fig. \ref{pose_graph_structure}.

\section{Experiments} \label{Experiments}
We use a Jackal robot equipped with a UWB tag (MDEK1001) and a LiDAR (Ouster1) to validate the performance of the proposed simultaneous calibration and localization framework. In the experiments, we use four UWB anchors (MDEK1001), one tag, and a LiDAR to localize a moving robot in a real indoor 3D environment. The anchors with different known heights are placed in the different corners of the environment. One anchor is set to be the initial anchor, one other anchor is set to be in the x or y direction of the initial anchor, and the rest of the anchors are with unknown positions. The robot is controlled to move along a designed trajectory with large spans in both the x and y directions. We initialize the positions of the UWB anchors using a mobile device communicating with the anchors with Bluetooth by trilateration described in Section \ref{selfCalibrationAnchor}, and then the initial position is used to calculate a rough position of the robot. A LiDAR odometry described in Section \ref{LiDARodometry} is used to estimate the relative poses of the robot, which are used to fuse with the rough positions.
\subsection{Sensor fusion and robot localization}
\begin{figure}[thpb]
      \centering
      \includegraphics[width=3.4in]{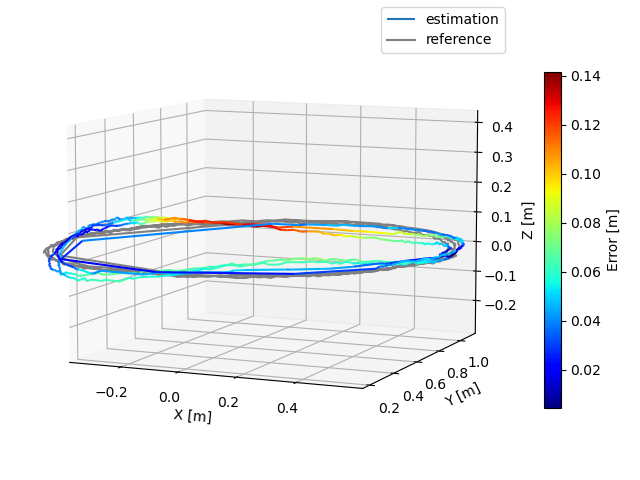}
      \caption{Result of the robot localization. The reference trajectory is provided by the LiDAR odometry during the same period of the calibration process. The estimation trajectory of the robot is calculated with the proposed FGO framework. The estimation trajectory is transformed into a reference frame to calculate the absolute distance error of each pose in the trajectory.}
      \label{robot_trajectory}
\end{figure}

The robot trajectories are shown in Fig. \ref{robot_trajectory}. We use the output of the LiDAR odometry as the reference and calculate the error of the global trajectory of the robot. We also calculate the error of the estimation only using UWB (UWB-only). The results are given in TABLE \ref{robot_trajectory_result}. The proposed FGO framework exhibits remarkable precision, achieving an accuracy of 0.0664 meters when evaluated using the Root Mean Square Error (RMSE). In contrast, the UWB-only approach yields an accuracy of 0.2397 meters, accompanied by a larger maximum error of 0.5949 meters. These results unequivocally underscore the efficacy and superiority of the proposed framework.

\begin{table}[ht]
\caption{Results of robot localization}
\centering
\label{robot_trajectory_result}
\begin{center}
\begin{tabular}{|c||c|c|c|}
\hline
Methods     & RMSE (m) & Max (m)   & Min (m) \\ \hline
UWB-only   & 0.2397     &   0.5949    &  0.0101 \\ \hline
Proposed   & 0.0664     &   0.1555    &  0.0060 \\
\hline
\end{tabular}
\end{center}
\end{table}
 
\subsection{Anchor calibration}
We first conducted 35 iterations using the self-calibration scheme outlined in Section \ref{selfCalibrationAnchor} (referred to as SCs), which requires approximately 1 hour to complete. We compute the mean and standard variance of these anchor positions. After self-calibration, we attach the anchors to different tripods with known heights in the same places and select one group of these anchor positions (a total of 35 groups) as the initial positions for our proposed framework. The selected initial positions of four anchors are (0,0,2.08), (0.12,2.93,0.98), (4.12,0,0.78), and (4.32,3.02,0.30). One other group of anchor positions  $(x_i,y_i,z_i), i=1,...,4$ measured by a ranging finder are (0,0,2.08), (0,3.00,0.98), (4.20,0,0.78), and (4.20,3.00,0.30), which can be used as a reference.

\begin{figure}[thpb]
      \centering
      \includegraphics[width=3.4in]{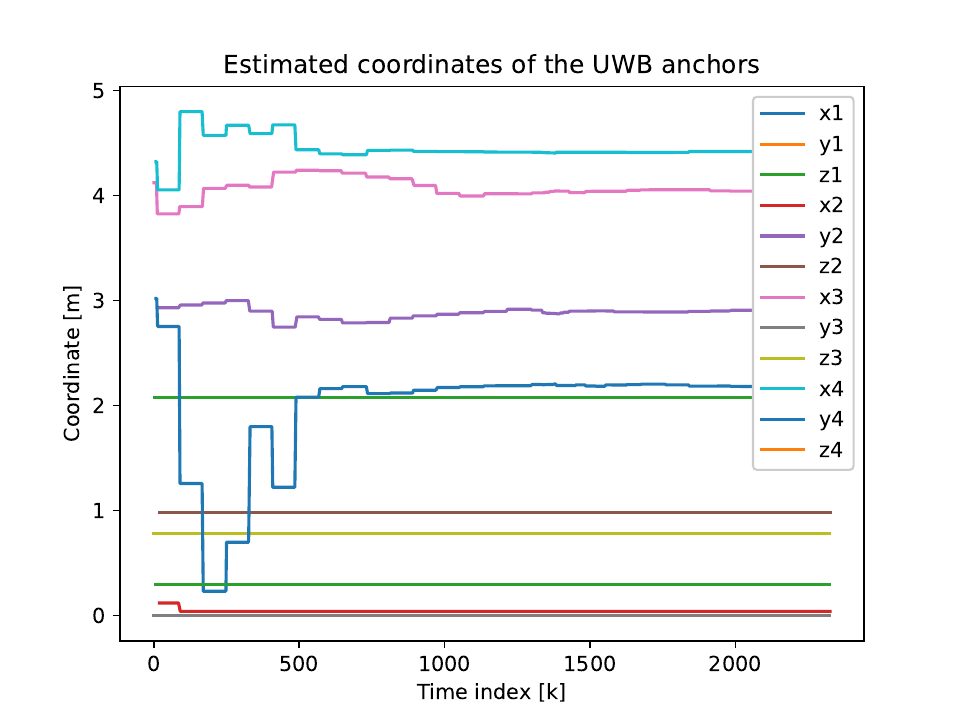}
      \caption{Result of the UWB-anchor calibration. The coordinates of four UWB anchors are calculated with the proposed framework. The positions of the anchors are estimated using the ranging measurements between the robot and the corresponding anchors. Some of the coordinates are set to be constant to set up a fixed human-known frame.}
      \label{anchor_result}
\end{figure}

We compare the final output of the proposed soft sensor with the mean of SCs. The detailed results of the variables can be found in TABLE \ref{anchor_compeare_result}. Additionally, the converging outcomes of the anchor position estimation are visually depicted in Fig. \ref{anchor_result}. The UWB ranging frequency is around 10 Hz. As can be found from Fig. \ref{anchor_result}, the calibration result converges after receiving $k=1000$ ranging readings, which means it only takes approximately 25 seconds to estimate the anchor positions, while also localizing the robot. These findings demonstrate the effectiveness and accuracy of our proposed approach. 

Note that the UWB-anchor calibration problem is solved simultaneously with the robot localization problem in the same FGO framework. Some estimations of the anchors (such as $y_2$) achieve higher accuracy than SCs. One can find that the error of $y_4$ reaches 0.8174 meters (less than 1 meter, the maximum error of SCs described in \cite{jin2021auto}), which indicates a large-ranging error and the unstable frame transformation between the frame ${U}$ and the frame ${M}$ at the beginning of the calibration will affect the estimation accuracy of some anchors. Some solutions for this problem include improving the accuracy of the UWB ranging measurements, considering the ill-conditioning of the calibration problem caused by the anchor layout and improving the calibration accuracy of the first few seconds by starting the calibration process after getting a stable frame transformation.

\begin{table}[ht]
\caption{Result of anchor calibration}
\label{anchor_compeare_result}
\begin{center}
\begin{tabular}{|c||c||cc|c|}
\hline
\multirow{2}{*}{Coordinates}  & Proposed    & \multicolumn{2}{c|}{SCs}  & Ranging finder  \\ \cline{2-5} 
                             & Final (m)       & \multicolumn{1}{c|}{Mean (m)}       & Std (m) & Manual (m)\\ \hline
$x_2$                        & 0.0398       & \multicolumn{1}{c|}{0.0209}     & 0.1956 & 0.12\\ \hline
$y_2$                        & 2.8950       & \multicolumn{1}{c|}{2.7317}     & 0.2634 & 2.93   \\ \hline
$x_3$                        & 4.0480       & \multicolumn{1}{c|}{4.1249}     & 0.0953 & 4.12   \\ \hline
$x_4$                        & 4.4140       & \multicolumn{1}{c|}{4.2574}     & 0.1293 & 4.32   \\ \hline
$y_4$                        & 2.1836       & \multicolumn{1}{c|}{2.9940}     & 0.0443 & 3.00   \\ \hline
\end{tabular}
\end{center}
\end{table}

\section{CONCLUSIONS} \label{CONCLUSIONS}
We propose a novel factor graph optimization framework to solve Ultra-WideBand anchor calibration problems and robot localization problems simultaneously and automatically, which saves time for UWB network deployment and can avoid artificial errors. The proposed method enables the creation of a soft sensor providing the position information of the anchors in a UWB network suitable for emergencies or situations without special calibration tools. The proposed soft sensor only requires UWB and LiDAR measurements measured on a moving robot. Moreover, the proposed method is applicable for large and extendable UWB networks as the only required information to calibrate a new anchor is the ranging measurements between the moving UWB tag on the robot and this new anchor. With the proposed framework, one can get a robust position estimation of the robot, even if the reading of the UWB ranging or the LiDAR is interrupted for a while, which means that the position estimation process will continue with only one sensor reading and ensure an acceptable estimation accuracy during the reading interruption of another sensor (UWB anchors or LiDAR). The position estimation process will be improved and the data fusion process be carried out after the readings return. The LiDAR can be replaced with any other odometry sensors, such as visual odometry and traditional odometry, IMU, and even GPS, and so on. With the help of the proposed framework, the accuracy and robustness of the robot pose estimation can be improved. From the experiment result, the calibration problem can be solved within 30 seconds by the proposed soft sensor and simultaneously, the robot localization accuracy achieves a sufficiently high level through sensor fusion of LiDAR SLAM and UWB. Future improvements of the proposed framework include improving the UWB ranging measurement by filtering, recognizing the NLOS environment or detecting multiple paths, improving the calibration accuracy of the first few seconds by starting the calibration process after getting a stable frame transformation, using LiDAR SLAM with loop closure detection and dynamic environment recognition, and combining two factor graphs into a cohesive framework.

\section*{ACKNOWLEDGMENT}
The authors would like to thank Bayu Jayawardhana for the UWB sensors and Simon Busman for his help in the experiments.


\end{document}